\documentclass[conference]{IEEEtran}
\IEEEoverridecommandlockouts
\usepackage{cite}
\usepackage{amsmath,amssymb}
\usepackage{graphicx}
\usepackage{booktabs}
\usepackage{multirow}
\usepackage{placeins}
\usepackage{stfloats}
\usepackage{balance}
\usepackage{url}
\usepackage{microtype}
\usepackage{xspace}
\newcommand{\TargetExposure}{\textsc{TargetExposure}@10\xspace}
\newcommand{\AnyExposure}{\textsc{AnyPoisonExposure}@10\xspace}
\newcommand{\SafeR}{\textsc{SafeRetrievalReliability}@10\xspace}
\begin{document}
\title{Reliability Stress Testing of RAG Retrieval under Query-Relevant Corpus Poisoning}
\author{
\IEEEauthorblockN{Janet Sung}
\IEEEauthorblockA{\textit{Google LLC}\\
Mountain View, USA\\
janetsung@google.com}
\and
\IEEEauthorblockN{Kainan Zhou}
\IEEEauthorblockA{\textit{Google LLC}\\
Mountain View, USA\\
zhoumark@google.com}
\and
\IEEEauthorblockN{Gangzhen Qian}
\IEEEauthorblockA{\textit{Google LLC}\\
Mountain View, USA\\
irisqian@google.com}
\and
\IEEEauthorblockN{Hang Xiao}
\IEEEauthorblockA{\textit{Fortinet, Inc.}\\
Sunnyvale, USA\\
hangxiao.pisces@gmail.com}
\and
\IEEEauthorblockN{Lu Yi}
\IEEEauthorblockA{\textit{Google LLC}\\
Mountain View, USA\\
annabelyi@google.com}
}

\maketitle

\begin{abstract}
External text can affect a retrieval-augmented generation (RAG) system only after the retriever admits it to the model context. We stress this admission step while varying document content, attacker query information, per-target poison budget, and contamination scope. Across 623 Benchmarking Information Retrieval (BEIR) queries from SciFact and NFCorpus, five query-targeted documents per query reduce classical-retriever nDCG@10 by 40.2--61.5\% in isolated-target stress. Full-query anchoring explains most Best Matching 25 (BM25) degradation, whereas appended text also shifts latent semantic analysis (LSA) and Hybrid rankings. A pretrained BGE-small validation reaches 100\% target exposure for query-only $P=5$ on both datasets; adding the injection payload changes SciFact nDCG only from 0.269 to 0.271 and lowers NFCorpus exposure slightly to 98.8\%, indicating that query anchoring remains the dominant retrieval-side factor for this dense model. Query length materially changes the interpretation of partial-information attacks: on 162 long SciFact queries, two random terms cover 19.9\% of eligible terms and still yield 39.5\% Hybrid target exposure. Multi-target contamination adds cross-target interference. Overlap-based admission can remove full-query poisons but may sharply reduce clean retrieval utility, and maximal marginal relevance (MMR) leaves exposure unchanged under the tested configuration. The study measures retrieval-layer failures, not downstream instruction following.
\end{abstract}

\begin{IEEEkeywords}
retrieval-augmented generation, reliability, prompt injection, corpus poisoning, information retrieval, admission control
\end{IEEEkeywords}

\section{Introduction}
A retrieval-augmented generation (RAG) system selects evidence from an external corpus before passing it to a language model~\cite{lewis2020rag}. Some indexed documents may be controlled by third parties. In \emph{corpus poisoning}, an attacker inserts adversarial material into these retrievable sources; a document resembling the user's query may enter the top-$k$ context and displace legitimate evidence before any malicious instruction is executed.

For example, an attacker can publish a page repeating terms from a scientific question and append misleading text. Although unable to change the query or retriever, the attacker can influence which evidence is admitted. We measure that retrieval boundary, not downstream instruction following.

Prior work studies knowledge corruption, retrieval jamming, optimized triggers, and indirect prompt injection~\cite{zou2025poisonedrag,shafran2025machine,chang2026retrievalbarrier,yi2025bipia}. We isolate retrieval-stage effects that are difficult to distinguish when document content, query knowledge, poison budget, and contamination scope change together.

We organize the evaluation around three research questions:
\begin{enumerate}
\item[\textbf{RQ1.}] What makes a poisoned document retrievable?
\item[\textbf{RQ2.}] How does retrieval failure scale with attacker query information and contamination scope?
\item[\textbf{RQ3.}] What does admission control trade for lower exposure?
\end{enumerate}
Public Benchmarking Information Retrieval (BEIR) labels and Benchmark for Indirect Prompt Injection Attacks (BIPIA) payload text support matched content controls, isolated and multi-target contamination, graded query-information conditions, and post-filter measurements.

We measure ranking quality, poison exposure, retained evidence, safe retrieval reliability, and latency. Query-information analysis accounts for effective query length: two or four terms disclose different fractions of short and long queries. The experiments distinguish query anchoring, cross-target contamination, and the evidence cost of admission rules.

\section{Related Work}
\subsection{Indirect Injection} BIPIA benchmarks indirect prompt injection and associated defenses~\cite{yi2025bipia}; InjecAgent extends indirect injection to tool-integrated agents~\cite{zhan2024injecagent}. Those studies evaluate downstream instruction following. Here, BIPIA payloads are used as standardized appended content, and the measured endpoint is retrieval admission.

\subsection{Retrieval Poisoning and Jamming} PoisonedRAG jointly optimizes retrieval and generation objectives~\cite{zou2025poisonedrag}, and AgentPoison targets agent memory and RAG knowledge bases with optimized triggers~\cite{chen2024agentpoison}. Machine Against the RAG shows that retrieval disruption need not rely on instruction semantics~\cite{shafran2025machine}. Chang \emph{et al.} characterize a retrieval barrier and optimize triggers across modern embedding models~\cite{chang2026retrievalbarrier}. Here, query information, per-target budget, and contamination scope are varied directly so their retrieval effects can be measured separately.

\subsection{Reliability and Admission Defenses} ReliabilityRAG uses document-reliability signals and provides stronger corruption defenses and robustness guarantees~\cite{shen2025reliabilityrag}. We use an interpretable overlap rule to measure the trade-off between lower poison exposure and loss of legitimate evidence; maximal marginal relevance (MMR)~\cite{carbonell1998mmr} provides a diversity-reranking comparison. The measurement distinction is that the closest prior work primarily studies downstream attack effects or optimized retrieval disruption, whereas our evaluation makes query information and contamination scope explicit stress axes and reports poison exposure together with retained evidence and ranking utility.

\section{Methodology and System Model}
\begin{figure*}[!t]
\centering
\includegraphics[width=0.80\textwidth]{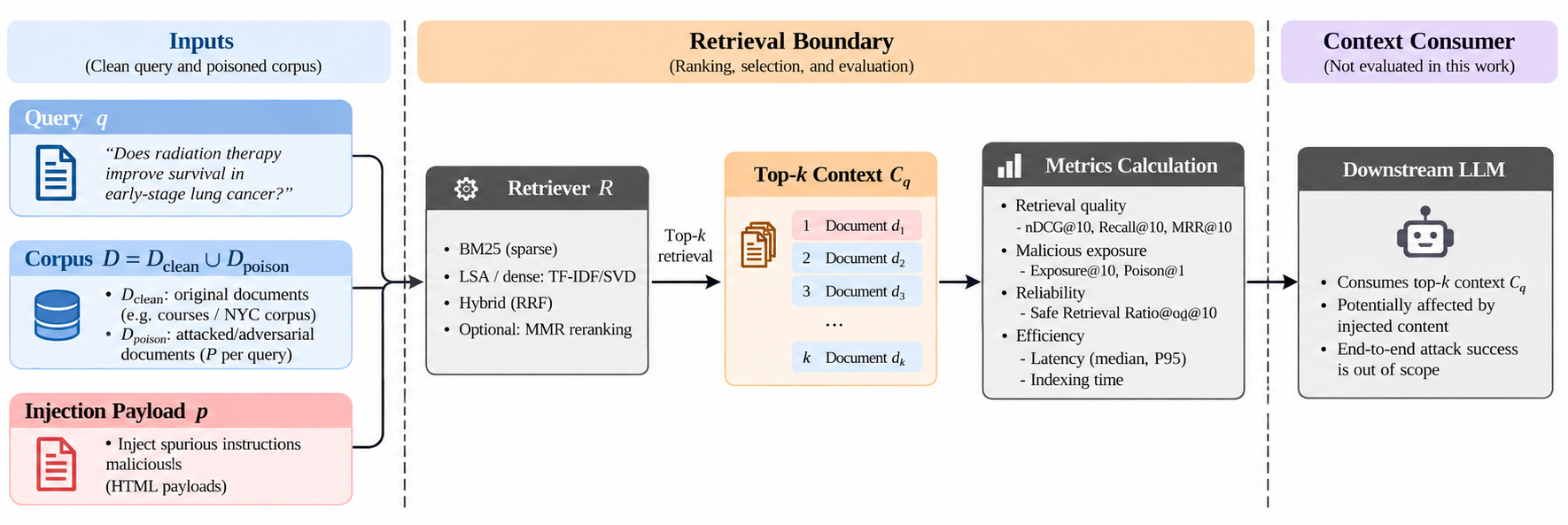}
\caption{System model and retrieval measurement boundary. Attacker-controlled documents enter the corpus; retrieval and optional admission determine the final top-$k$ set. Downstream instruction following is not evaluated.}
\label{fig:framework}
\end{figure*}

Figure~\ref{fig:framework} shows the system model: attacker-controlled documents enter a fixed corpus, retrieval ranks the combined collection, and optional reranking or filtering determines the admitted candidates. We independently vary content, query knowledge, and contamination scope, then measure exposure and utility. Downstream generation is excluded.
\subsection{Threat Model and Stress Axes}
Let $\mathcal{D}$ be a clean corpus, $q$ a benchmark query, $\mathcal{R}_q$ its judged-relevant documents, and $\mathcal{A}_q(P)$ a set of $P$ query-targeted poisoned documents. For a set of contaminated targets $\mathcal{G}$, the poison corpus is
\begin{equation}
\mathcal{A}=\bigcup_{q\in\mathcal{G}}\mathcal{A}_q(P),\qquad
\mathcal{T}_k(q)=\operatorname{TopK}(q,\mathcal{D}\cup\mathcal{A}).
\end{equation}
The attacker can publish indexed content but cannot modify the benchmark query, clean documents, relevance labels, retriever parameters, or $k$. We vary three stress axes. \emph{Per-target budget} is $P\in\{1,3,5\}$. \emph{Contamination scope} ranges from one target at a time to all evaluation targets, and is reported with both the number of poison targets/documents and global corpus fraction. \emph{Query information} ranges from no query terms, through two or four target-query terms, to the full query. Eligible terms are unique lowercase alphanumeric tokens of length at least three, after stop-word removal and restriction to the clean BM25 vocabulary; if fewer than the requested two or four terms remain, all eligible terms are used. High-inverse-document-frequency (IDF) partial anchors are an informed/oracle-like condition because IDF is computed from the clean corpus; deterministic random-query-term anchors use the same term budget without corpus-statistics knowledge. We report the actual fraction of eligible query terms revealed so that a fixed term budget is not mistaken for a fixed information fraction. These conditions are stress assumptions, not estimates of real-world attack frequency. Thus, query information and contamination scope are controlled directly rather than left implicit in a single attack-strength setting.

\subsection{Retrieval Reliability Metrics}
Target exposure is
\begin{equation}
E_q=\mathbf{1}[\mathcal{T}_{10}(q)\cap\mathcal{A}_q(P)\neq\varnothing],
\end{equation}
and \TargetExposure is its empirical mean. \AnyExposure counts any poisoned document, including one targeted to a different query; we also report target and any-poison counts in the top ten. Evidence availability is $H_q=\mathbf{1}[\mathcal{T}_{10}(q)\cap\mathcal{R}_q\neq\varnothing]$. Safe retrieval reliability is
\begin{equation}
S_q=\mathbf{1}[H_q=1\land \mathcal{T}_{10}(q)\cap\mathcal{A}=\varnothing].
\end{equation}
Standard nDCG@10 and Recall@10 measure ranking utility. $S_q$ is a strict component-level conjunction---evidence present and zero poison admitted---rather than a deployment safety probability. We therefore also decompose attacked queries into four observable states: evidence+poison, evidence+no-poison, no-evidence+poison, and no-evidence+no-poison. Together, these metrics distinguish poison admission, evidence loss, cross-target interference, and ranking degradation.

\section{Experimental Design}
We instantiate the system model using public retrieval benchmarks, matched content controls, and isolated-target and shared-corpus stress. Clean-corpus retrieval statistics stay fixed so the comparisons isolate the effect of added documents.

\subsection{Datasets and Retrieval}
We use the BEIR test splits of SciFact (5,183 documents, 300 judged queries) and NFCorpus (3,633 documents, 323 judged queries)~\cite{thakur2021beir}. Retrieval uses Best Matching 25 (BM25)~\cite{robertson2009bm25}, 64-dimensional latent semantic analysis (LSA)~\cite{deerwester1990lsa}, and reciprocal-rank fusion (Hybrid)~\cite{cormack2009rrf}. Hybrid uses RRF constant 60, fuses the top 100 BM25 and LSA candidates, and retains 50 candidates before the final top-10 stage. For external validation, we also evaluate the pretrained 33M-parameter \texttt{BAAI/bge-small-en-v1.5} bi-encoder~\cite{xiao2023cpack} with normalized embeddings and cosine similarity, without task-specific training.

For isolated-target experiments, LSA is fitted once on the clean corpus and injected documents are transformed into that frozen space. BM25 likewise uses vocabulary, IDF, and average document-length statistics fitted to the clean corpus; poison documents are scored under those frozen statistics. Multi-target causal controls use the same clean-corpus statistics, so differences reflect added documents rather than retraining the representation.

\subsection{Content Controls and Stress Regimes}
For each target query and $P\in\{1,3,5\}$, query-only poisons contain the target query without appended text; query+payload poisons append a BIPIA payload; fixed-neutral controls append deterministic benign filler of matched length. Payloads are sampled deterministically with replacement from four BIPIA text-attack categories (Base Encoding, Reverse Text, Emoji Substitution, and Language Translation) using seed 2026 and a query-specific hash. A stronger corpus-derived neutral condition uses five deterministic non-relevant clean-corpus snippets per target: each source has zero BM25 query score, is excluded from judged-relevant documents, and is truncated to the payload token length. This control estimates perturbation from appended corpus-like text without claiming perfect lexical, topical, or semantic matching to the payload.

In \emph{isolated-target} stress, each query is evaluated against the clean corpus plus only its own $P$ poisons. At $P=5$, five targeted documents represent 0.10\% of SciFact and 0.14\% of NFCorpus. In \emph{multi-target} stress, every evaluation query contributes $P$ poisons to the same corpus; at $P=5$, this yields 1,500 SciFact poisons (22.45\%) and 1,615 NFCorpus poisons (30.77\%). For the attacker-information axis at isolated $P=5$, payload-only provides no query terms; two- and four-term conditions use either deterministic random eligible query terms or the highest-IDF eligible terms; full-query stress includes the complete query. Random-term selection uses a fixed query-specific SHA-256-derived seed; all term-budget conditions reuse the same per-query payload construction.

\subsection{Admission Controls, Statistics, and Latency}
MMR reranks the same 50-candidate pool with $\lambda=0.7$, chooses the highest-relevance candidate first, and then trades normalized relevance against LSA-vector redundancy. Admission control uses IDF-weighted unique-query-term coverage $C(q,d)$ and rejects candidates with $C(q,d)\geq\tau$, where $\tau\in\{0.25,0.40,0.55,0.70,0.85,0.95\}$. Filtering scans the ranked 50-candidate list and continues downward until ten candidates pass or the pool is exhausted; shortages are reported explicitly. For clean, four-term, and full-query conditions, each threshold is evaluated on final top-10 nDCG, Recall, evidence availability, safe retrieval reliability, target exposure, and the fraction of queries with fewer than ten admitted candidates. We report the full frontier rather than selecting a test-set-optimal threshold.

Metrics are computed per query. Nonparametric bootstrap 95\% confidence intervals use queries as the resampling unit; paired intervals compare query+payload with query-only and with corpus-derived neutral content. Latency uses 500 repeated $P=5$ measurements per retriever/defense pair on an AMD EPYC 9V74 allocation with five visible CPU cores under Python 3.13.5, reporting absolute median and P95 milliseconds.

\section{Results}
We first examine content controls for RQ1, then attacker information and contamination scope for RQ2, and finally admission trade-offs for RQ3. Each analysis measures the retrieved candidates rather than a generated answer.
\subsection{RQ1: What makes a poisoned document retrievable?}
\begin{figure*}[!t]
\centering
\includegraphics[width=0.78\textwidth]{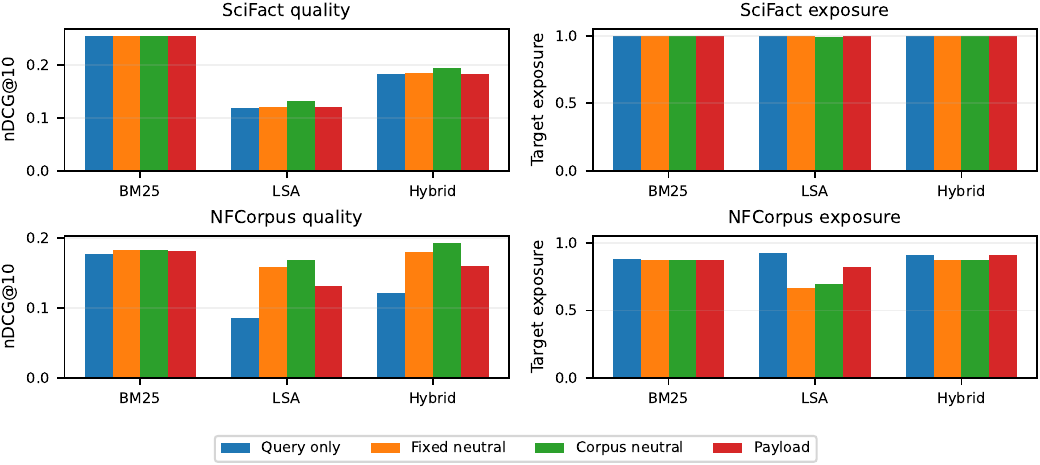}
\caption{Isolated-target $P=5$ content controls for nDCG@10 and target exposure. Table~\ref{tab:causalci} reports the paired confidence intervals.}
\label{fig:controls}
\end{figure*}
Figure~\ref{fig:controls} compares isolated-target $P=5$ query-only, fixed-neutral, corpus-derived-neutral, and payload conditions. On SciFact, query-only and query+payload nDCG are identical for BM25 and differ by only +0.0014 for LSA and +0.0003 for Hybrid. On NFCorpus, payload-minus-query-only differences are +0.0039 for BM25, +0.0450 for LSA, and +0.0382 for Hybrid, indicating stronger sensitivity of semantic representations to appended text.

The corpus-derived neutral control separates query anchoring from appended-content identity more carefully. For BM25, payload-minus-neutral nDCG is 0 on SciFact and $-0.0004$ on NFCorpus (95\% CI $[-0.0013,0.0000]$). For LSA/Hybrid, the corresponding differences are $-0.0117/-0.0108$ on SciFact and $-0.0370/-0.0367$ on NFCorpus, with paired intervals excluding zero. NFCorpus target exposure is also 13.3 percentage points higher for LSA and 3.4 points higher for Hybrid under payload than corpus-neutral text. On BM25, the query anchor accounts for nearly all of the observed degradation. LSA and Hybrid are more sensitive to the appended text, especially on NFCorpus. Because the neutral additions also change these rankings, the experiment supports an appended-content effect rather than a payload-specific retrieval mechanism.

Table~\ref{tab:causalci} gives paired bootstrap intervals for the same comparisons. SciFact Hybrid changes by only $+0.0003$ relative to query-only (95\% CI $[-0.0002,0.0010]$), whereas NFCorpus LSA and Hybrid shift by $+0.0450$ and $+0.0382$. Against corpus-derived neutral text, the LSA/Hybrid differences are negative on both datasets and their intervals exclude zero, showing that the identity of the appended text can change semantic rankings even with the same query anchor.

\begin{table}[t]
\caption{Paired $\Delta$ nDCG@10 at isolated $P=5$ with 95\% bootstrap CIs.}
\label{tab:causalci}
\centering
\scriptsize
\resizebox{\columnwidth}{!}{%
\begin{tabular}{llcc}
\toprule
Data & Retr. & $\Delta$ vs. query-only & $\Delta$ vs. corpus-neutral \\
\midrule
SciFact & BM25 & $+.0000[+.0000,+.0000]$ & $+.0000[+.0000,+.0000]$ \\
SciFact & LSA & $+.0014[+.0000,+.0038]$ & $-.0117[-.0159,-.0080]$ \\
SciFact & Hybrid & $+.0003[-.0002,+.0010]$ & $-.0108[-.0174,-.0051]$ \\
NFCorpus & BM25 & $+.0039[+.0015,+.0069]$ & $-.0004[-.0013,+.0000]$ \\
NFCorpus & LSA & $+.0450[+.0329,+.0586]$ & $-.0370[-.0458,-.0292]$ \\
NFCorpus & Hybrid & $+.0382[+.0278,+.0494]$ & $-.0367[-.0445,-.0292]$ \\
\bottomrule
\end{tabular}}
\end{table}

\subsection{RQ2: How does failure scale with attacker information and contamination scope?}
\begin{table*}[!t]
\caption{Core retrieval and reliability metrics under clean, isolated $P=5$, and multi-target $P=5$ query+payload stress.}
\label{tab:closure}
\centering
\scriptsize
\resizebox{0.91\textwidth}{!}{%
\begin{tabular}{lllrrrrrrrr}
\toprule
Data & Retr. & Condition & nDCG & Recall & TargetExp & AnyExp & Evid. & SafeR & \#P & Poison\% \\
\midrule
\multirow{9}{*}{SciFact}
& BM25 & Clean & .662 & .781 & .000 & .000 & .800 & .800 & .00 & .00 \\
& BM25 & Isolated $P=5$ & .255 & .718 & 1.000 & 1.000 & .740 & .000 & 5.00 & .10 \\
& BM25 & Multi $P=5$ & .188 & .529 & 1.000 & 1.000 & .550 & .000 & 6.22 & 22.45 \\
& LSA & Clean & .295 & .442 & .000 & .000 & .453 & .453 & .00 & .00 \\
& LSA & Isolated $P=5$ & .119 & .347 & 1.000 & 1.000 & .367 & .000 & 4.97 & .10 \\
& LSA & Multi $P=5$ & .085 & .244 & 1.000 & 1.000 & .257 & .000 & 7.16 & 22.45 \\
& Hybrid & Clean & .465 & .649 & .000 & .000 & .663 & .663 & .00 & .00 \\
& Hybrid & Isolated $P=5$ & .183 & .530 & 1.000 & 1.000 & .547 & .000 & 5.00 & .10 \\
& Hybrid & Multi $P=5$ & .138 & .398 & 1.000 & 1.000 & .410 & .000 & 6.34 & 22.45 \\
\midrule
\multirow{9}{*}{NFCorpus}
& BM25 & Clean & .308 & .146 & .000 & .000 & .678 & .678 & .00 & .00 \\
& BM25 & Isolated $P=5$ & .181 & .118 & .876 & .876 & .619 & .037 & 4.28 & .14 \\
& BM25 & Multi $P=5$ & .153 & .108 & .972 & .972 & .548 & .022 & 5.49 & 30.77 \\
& LSA & Clean & .221 & .109 & .000 & .000 & .563 & .563 & .00 & .00 \\
& LSA & Isolated $P=5$ & .131 & .090 & .824 & .824 & .498 & .071 & 3.64 & .14 \\
& LSA & Multi $P=5$ & .091 & .066 & .805 & .966 & .328 & .022 & 7.23 & 30.77 \\
& Hybrid & Clean & .267 & .129 & .000 & .000 & .622 & .622 & .00 & .00 \\
& Hybrid & Isolated $P=5$ & .160 & .099 & .910 & .910 & .551 & .006 & 4.22 & .14 \\
& Hybrid & Multi $P=5$ & .138 & .091 & 1.000 & 1.000 & .520 & .000 & 6.22 & 30.77 \\
\bottomrule
\end{tabular}}
\end{table*}
Table~\ref{tab:closure} shows that contamination scope changes the failure mode, not merely its magnitude. Under isolated $P=5$, five query-targeted documents constitute only 0.10--0.14\% of the corpus, yet Hybrid nDCG falls from 0.465 to 0.183 on SciFact and from 0.267 to 0.160 on NFCorpus; \SafeR falls from 0.663 to 0 and from 0.622 to 0.006, respectively. The small global fractions need to be read together with the strong per-query targeting: this condition measures concentrated query-aware contamination, not random corruption at the same corpus fraction.

\begin{table}[t]
\caption{Hybrid isolated full-query stress as the per-target budget $P$ increases.}
\label{tab:budget}
\centering
\scriptsize
\begin{tabular}{lrrrrr}
\toprule
Data & $P$ & nDCG & TargetExp & Evid. & SafeR \\
\midrule
\multirow{3}{*}{SciFact} & 1 & .328 & 1.000 & .647 & .000 \\
& 3 & .232 & 1.000 & .593 & .000 \\
& 5 & .183 & 1.000 & .547 & .000 \\
\midrule
\multirow{3}{*}{NFCorpus} & 1 & .230 & .889 & .607 & .025 \\
& 3 & .186 & .907 & .573 & .009 \\
& 5 & .160 & .910 & .551 & .006 \\
\bottomrule
\end{tabular}
\end{table}
Table~\ref{tab:budget} closes the $P\in\{1,3,5\}$ budget sweep for the Hybrid isolated setting. On SciFact, target exposure is already 1.0 at $P=1$, while additional poisons mainly displace evidence and reduce nDCG. On NFCorpus, exposure changes less than ranking quality as $P$ grows: nDCG falls from 0.230 at $P=1$ to 0.160 at $P=5$, while exposure moves from 0.889 to 0.910.

\begin{table}[t]
\caption{Hybrid isolated $P=5$ four-state decomposition (fractions of queries).}
\label{tab:fourstate}
\centering
\scriptsize
\resizebox{\columnwidth}{!}{%
\begin{tabular}{lrrrr}
\toprule
Data & Evid.+Poison & Evid.+NoPoison & NoEvid.+Poison & NoEvid.+NoPoison \\
\midrule
SciFact & .547 & .000 & .453 & .000 \\
NFCorpus & .545 & .006 & .365 & .084 \\
\bottomrule
\end{tabular}}
\end{table}
Table~\ref{tab:fourstate} clarifies the strict SafeR definition. SciFact SafeR is zero because every attacked query admits poison, even though 54.7\% still retain judged evidence. NFCorpus likewise retains evidence alongside poison for 54.5\% of queries; only 0.6\% satisfy the stricter evidence-plus-zero-poison condition. SafeR therefore summarizes a component-level conjunction, not ``probability the RAG system answers incorrectly.''

Under multi-target $P=5$, all targets contribute poisons simultaneously, creating 22.45\% and 30.77\% corpus fractions. Cross-target interference becomes measurable. For NFCorpus LSA, target exposure is 0.805 while any-poison exposure is 0.966 and the mean top-10 poison count is 7.23. Target exposure alone misses a substantial part of the saturation regime. Reporting per-target $P$, contaminated-target count, corpus fraction, and any-poison exposure distinguishes concentrated targeting from broad corpus stress.

\begin{figure}[!t]
\centering
\includegraphics[width=0.80\columnwidth]{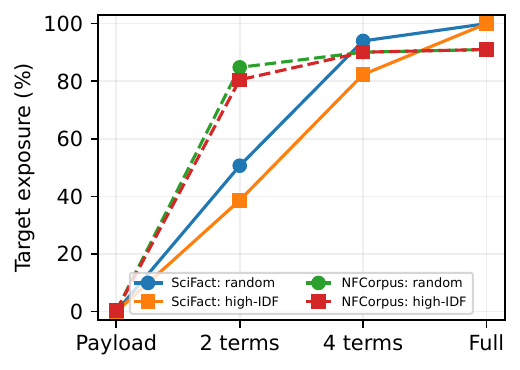}
\caption{Hybrid target exposure versus attacker query information at isolated $P=5$.}
\label{fig:attacker_info}
\end{figure}

Attacker query information forms a second stress axis, but a fixed term count is not a fixed information fraction. Figure~\ref{fig:attacker_info} reports the unconditioned Hybrid exposure curves; Table~\ref{tab:querycoverage} supplies the missing query-length context. NFCorpus has only 2.48 eligible terms on average (median 2): 57.6\% of its queries are fully covered by the two-term budget and 90.7\% by the four-term budget. Its high 2/4-term exposure therefore should not be read as evidence that a small fraction of query information is always sufficient.

\begin{table}[t]
\caption{Random-term information coverage and Hybrid exposure at isolated $P=5$. ``Long'' means at least eight eligible terms.}
\label{tab:querycoverage}
\centering
\scriptsize
\begin{tabular}{lrrrrr}
\toprule
Data/subset & $n$ & Mean elig. & Cov$_2$ & Exp$_2$ & Cov$_4$ / Exp$_4$ \\
\midrule
SciFact all & 300 & 8.33 & .278 & .507 & .553 / .940 \\
SciFact long & 162 & 10.63 & .199 & .395 & .397 / .895 \\
NFCorpus all & 323 & 2.48 & .805 & .848 & .974 / .901 \\
\bottomrule
\end{tabular}
\end{table}
The long-query subset gives a cleaner partial-information test. On 162 SciFact queries with at least eight eligible terms, two random terms reveal 19.9\% of eligible terms on average yet still produce 39.5\% target exposure; four terms reveal 39.7\% and produce 89.5\% exposure. High-IDF anchors remain an informed/oracle-like condition, not a necessary attacker capability. These results support sensitivity to partial query information on sufficiently long queries while exposing the short-query confound on NFCorpus.

\subsection{Pretrained dense-retriever validation}
We repeat the isolated $P=5$ clean/query-only/query+payload comparison with \texttt{BAAI/bge-small-en-v1.5}. Query-only contamination reaches 100\% target exposure on both datasets (Table~\ref{tab:bge}). Adding the payload leaves SciFact essentially unchanged and is less disruptive on NFCorpus, where exposure falls slightly to 0.988. This compact replication extends high query-targeted admission to a pretrained dense retriever while again separating retrieval relevance from malicious payload semantics.

\begin{table}[t]
\caption{BGE-small isolated $P=5$ validation. Each attacked cell is nDCG@10 / target exposure@10.}
\label{tab:bge}
\centering
\scriptsize
\begin{tabular}{lccc}
\toprule
Data & Clean nDCG & Query only & Query+payload \\
\midrule
SciFact & .713 & .269 / 1.000 & .271 / 1.000 \\
NFCorpus & .345 & .137 / 1.000 & .205 / .988 \\
\bottomrule
\end{tabular}
\end{table}

\subsection{RQ3: What does admission control trade for lower exposure?}
\begin{figure}[!t]
\centering
\includegraphics[width=0.80\columnwidth]{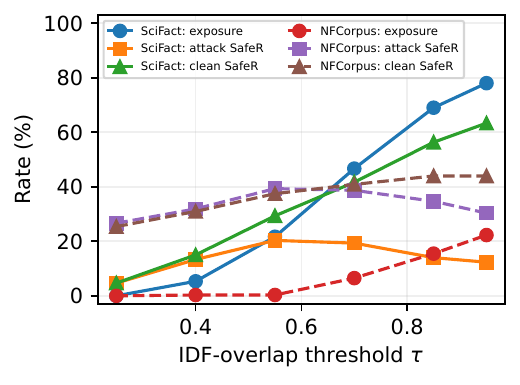}
\caption{Hybrid four-term overlap-admission frontier for exposure and safe retrieval reliability.}
\label{fig:admission_frontier}
\end{figure}
Figure~\ref{fig:admission_frontier} summarizes the overlap threshold security--utility frontier in final retrieval quality. For full-query Hybrid stress, $\tau=0.95$ removes target exposure on both datasets. This is an attack-matched rule: a document that repeats nearly all query terms is expected to have coverage near one, so full-query rejection at a 0.95 threshold is not itself evidence of a broadly effective detector. The informative result is the collateral cost. On SciFact, attack and clean nDCG under that threshold are both 0.440, close to the unfiltered clean 0.465. On NFCorpus, however, attack nDCG is 0.138 and clean nDCG is 0.134, far below the unfiltered clean 0.267; 3.4--4.0\% of queries also retain fewer than ten candidates. The four-term condition is more difficult to filter selectively: at $\tau=0.40$, target exposure falls to 5.3\% on SciFact and 0.3\% on NFCorpus, but clean nDCG collapses to 0.082 and 0.065. At $\tau=0.95$, clean nDCG recovers to 0.440/0.134 while four-term exposure rises to 78.0\%/22.3\%. Across the sweep, stronger lexical rejection reduces exposure but also removes legitimate evidence, and the trade-off differs sharply between the two datasets.

MMR closes a different question. Under isolated full-query $P=5$ Hybrid stress, target exposure remains 1.000 on SciFact and 0.910 on NFCorpus while nDCG changes from 0.183 to 0.234 and from 0.160 to 0.164; \SafeR remains 0 and 0.006. MMR therefore changes ranking utility without acting as an admission barrier here, while adding about 1.5~ms to median Hybrid latency on both datasets.

\section{Discussion}
\subsection{Document Content} Under full-query stress, BM25 is dominated by the query anchor, whereas LSA and Hybrid also respond to the appended text. The neutral controls show that this semantic-ranking shift is not specific to malicious instructions. The BGE-small replication extends the high-exposure result to a pretrained dense retriever, but its query+payload condition is no more damaging than query-only on SciFact and is less damaging on NFCorpus. This reinforces the distinction between query-relevant retrieval manipulation and downstream malicious-instruction semantics.

\subsection{Stress Intensity} Isolated $P=5$ combines a small global poison fraction with five documents written specifically for each evaluated query. Multi-target $P=5$ instead creates a broad saturation condition in which poison written for other targets also enters top-$k$. Query-information budgets require the same care: on short NFCorpus queries, two or four eligible terms often reveal nearly the whole query, whereas the long SciFact subset shows substantial exposure at 19.9--39.7\% mean coverage. A single poison rate or term count therefore hides materially different retrieval conditions.

\subsection{Admission Policy} Lower exposure does not necessarily improve final retrieval if the same rule removes judged evidence or leaves too few candidates. The overlap sweep has no stable cross-dataset operating point: thresholds that strongly suppress poison impose large clean-task costs on NFCorpus. Attack-side exposure and clean-task utility need to be evaluated together.

\section{Limitations}
The study ends at retrieval, so exposure is not equivalent to downstream attack success. Query-aware stressors are transparent rather than stealthy or optimized, and isolated low-fraction experiments still grant strong per-target targeting. Corpus-derived neutral snippets improve content matching but do not perfectly match payload term frequency, topic, or semantics; the analysis attributes effects to query anchoring versus appended-content identity rather than to malicious meaning itself. Fixed two/four-term budgets are especially difficult to interpret on short queries, so the long-query analysis is reported separately. High-IDF conditions additionally assume access to clean-corpus statistics. Pretrained validation is limited to one compact BGE model and the three core clean/query-only/query+payload conditions; larger embedding models, alternative dense retrievers, and production vector-search stacks remain outside scope. BEIR qrels are incomplete, so evidence availability and \SafeR inherit judgment incompleteness. Threshold curves are measurement frontiers rather than deployment recommendations.

\section{Conclusion}
The experiments provide direct answers to the three research questions. \textbf{RQ1:} Query anchoring explains most BM25 degradation; appended content can also shift LSA and Hybrid rankings without establishing a malicious-instruction-specific retrieval effect. The pretrained BGE-small check reaches 100\% target exposure for query-only poisoning on both datasets, with no additional payload-specific penalty in the tested conditions. \textbf{RQ2:} Two randomly selected terms yield 39.5\% Hybrid exposure on 162 long SciFact queries, whereas fixed term budgets can approximate full-query knowledge on short NFCorpus queries; broad contamination additionally causes cross-target interference. \textbf{RQ3:} Overlap admission suppresses full-query exposure but sacrifices clean ranking utility at dataset-dependent rates, and maximal marginal relevance (MMR) leaves exposure unchanged under the tested configuration. These are retrieval-stage findings, not downstream attack-success estimates.

\end{document}